\documentclass[]{bytedance_seed}

\usepackage[toc,page,header]{appendix}

\usepackage{minitoc}
\usepackage{microtype}
\usepackage{hyperref}
\usepackage{url}
\usepackage{booktabs}
\usepackage{graphicx}
\usepackage{lineno}
\usepackage{amsmath}
\usepackage{fancybox}
\usepackage{tikz}
\usepackage{wrapfig}
\usepackage{CJKutf8}

\title{Recitation over Reasoning: How Cutting-Edge Language Models Can Fail on Elementary School-Level Reasoning Problems?}

\author[1,2,*,\dagger]{Kai Yan}
\author[1,\dagger]{Yufei Xu}
\author[1]{Zhengyin Du}
\author[1]{Xuesong Yao}
\author[1]{Zheyu Wang}
\author[1]{Xiaowen Guo}
\author[1]{Jiecao Chen}

\affiliation[1]{ByteDance Seed}
\affiliation[2]{University of Illinois Urbana-Champaign}

\contribution[*]{Work done at ByteDance Seed}
\contribution[\dagger]{Corresponding authors}

\abstract{
The rapid escalation from elementary school-level to frontier problems of the difficulty for LLM benchmarks in recent years seems to bring us close enough to the ``last exam'' for LLMs to surpass humanity. However, is the LLMs' remarkable reasoning ability indeed coming from true intelligence by human standards, or are they actually reciting solutions witnessed during training at an Internet level? To study this problem, we propose RoR-Bench, a novel, multi-modal benchmark for detecting LLM's recitation behavior when asked simple reasoning problems but with conditions subtly shifted, and conduct empirical analysis on our benchmark. Surprisingly, we found existing cutting-edge LLMs unanimously exhibits extremely severe recitation behavior; by changing one phrase in the condition, top models such as OpenAI-o1 and DeepSeek-R1 can suffer $60\%$ performance loss on elementary school-level arithmetic and reasoning problems. Such findings are a wake-up call to the LLM community that compels us to reevaluate the true intelligence level of cutting-edge LLMs.
}

\date{\today}
\correspondence{Kai Yan at \email{kaiyan3@illinois.edu}, Yufei Xu at \email{xuyufei.123@bytedance.com}}
\checkdata[Dataset]{\url{https://huggingface.co/datasets/kaiyan289/RoR-Bench/tree/main}}

\begin{document}
\maketitle


\begin{figure}[ht]
    \centering
    \begin{minipage}[c]{0.4\linewidth}
        \centering
        \includegraphics[height=4.8cm]{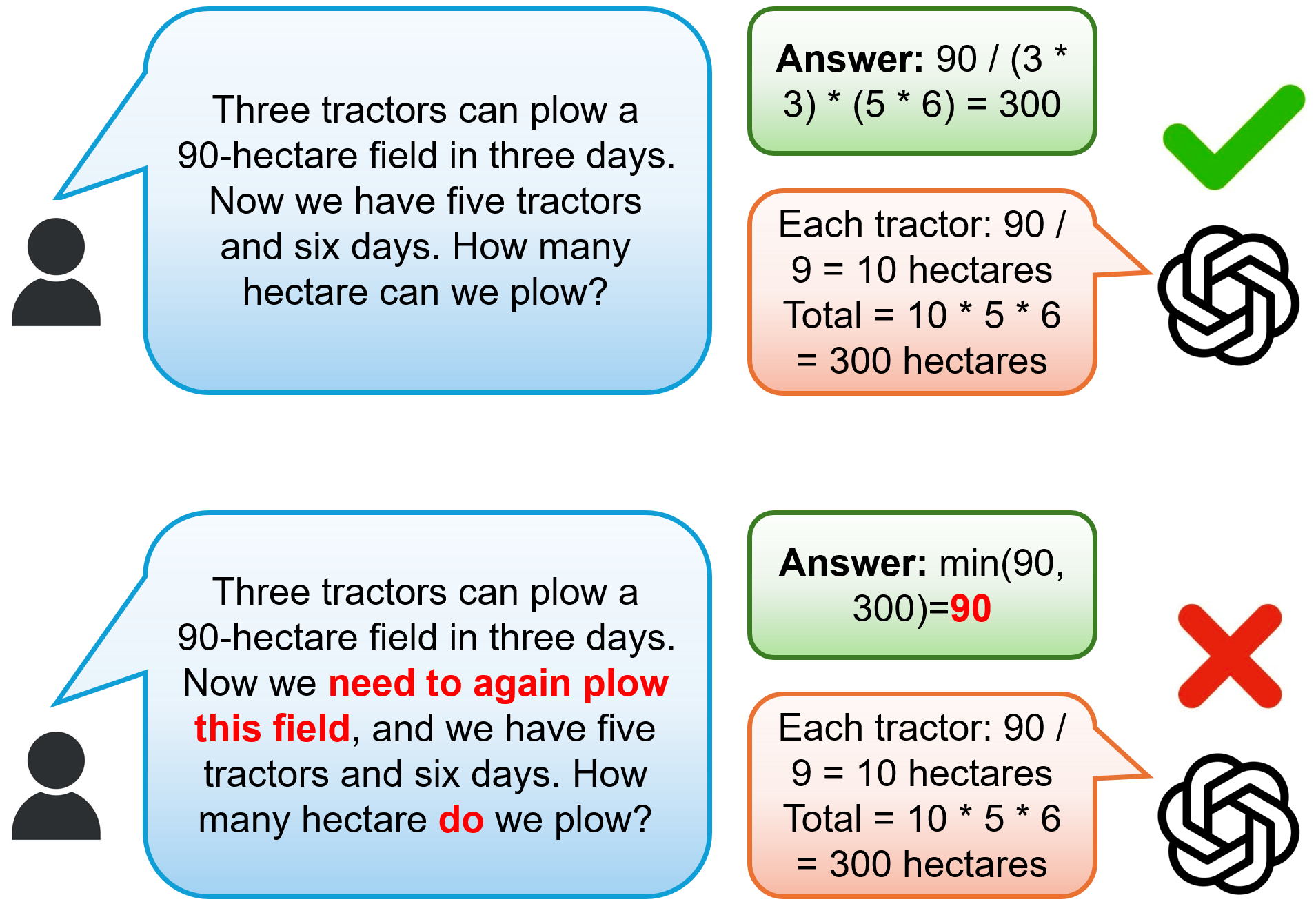}
        \caption*{a) Subtly changed condition}
    \end{minipage}
    \begin{minipage}[c]{0.4\linewidth}
        \centering
        \includegraphics[height=4.8cm]{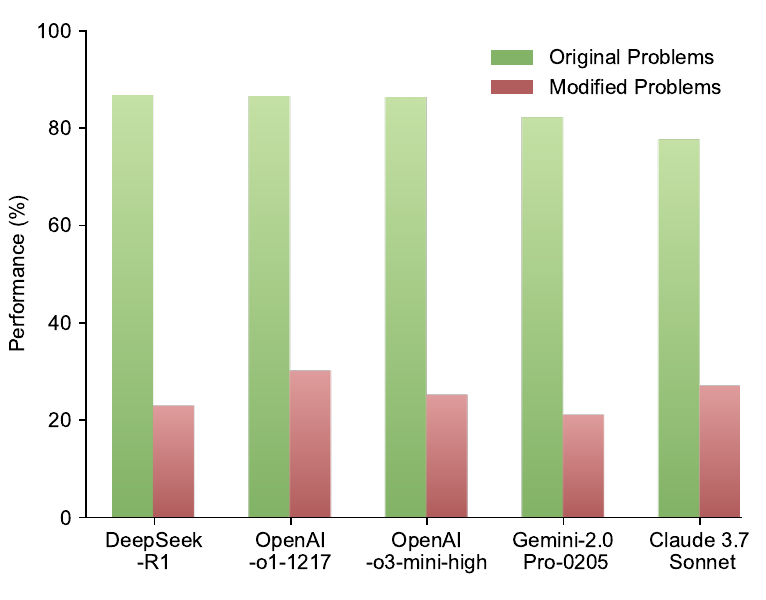}
        \caption*{b) Performance loss due to recitation}
    \end{minipage}
    \caption{Panel a) shows an example of how current cutting-edge LLMs, such as OpenAI-o1-1217~\citep{jaech2024openai}, OpenAI-o3~\citep{o3} and Gemini-Pro 2.5~\citep{comanici2025gemini} fails to address an elementary school-level math problem (see Appendix~\ref{sec:teaser_prompt} for links to the response) with subtle but crucial condition change, simply \textit{reciting} existing solution template; panel b) shows the performance loss of cutting-edge LLMs due to reciting solution templates regardless of shifted conditions on our benchmark, which is a staggering $\sim 60\%$ score gap on simple reasoning and math problems.}
    \label{fig:teaser}
\end{figure}

\section{Introduction}

Since the advent of GPT-3~\citep{brown2020language} and ChatGPT~\citep{ouyang2022training}, Large Language Models (LLMs) have sparked an unprecedented revolution of research paradigm and pushed forward task frontiers in almost every field of Artificial Intelligence (AI)~\citep{qin2024large, wang2024qwen2, ma2023eureka,zhou2023language}, as well as the whole science community~\citep{zhang2023huatuogpt,Abramson2024,zhang2024comprehensive}. By improving the training data~\citep{liu2024datasets,villalobos2024position}, scaling up parameter size~\citep{kaplan2020scaling,zhang2024scaling}, and incorporating long thinking process~\citep{jaech2024openai,guo2025deepseek}, LLMs finally come close enough to the ``last exam''~\citep{phan2025humanity} for Artificial General Intelligence (AGI) to surpass humanity. 

Despite the huge success of LLMs, however, researchers have not fully understood the underlying mechanism for LLM's ``emerging''~\citep{wei2022emergent,arora2023theory} intelligence via current engineering~\citep{dubey2024llama,guo2025deepseek} advances. While there have been many efforts from the researchers to theoretically guarantee LLMs' intelligence level~\citep{akyurek2022learning, bhargava2023s, zekri2024large} and rapid escalations in the difficulty of solvable math and science competition problems from elementary school~\citep{cobbe2021training} to research level~\citep{phan2025humanity}, there have also been recent concerns on LLMs are still struggling with real-world problems~\citep{wang2024gta}, even those which are not so difficult for humans~\citep{mirzadeh2024gsm,zhou2024larger}. Such works indicates that a cloud still exists upon the great monument of reasoning for LLMs, which questions the actual intelligence level of LLMs in reasoning problems and again brought the concern of ``stochastic parrots''~\citep{bender2021dangers} back to the table.

To better illustrate the existence of such cloud, here we examine a simple, GSM-8K~\citep{cobbe2021training} level math problem as an example in Fig.~\ref{fig:teaser}. Despite the simplicity of the problem, however, cutting-edge models such as OpenAI o1~\citep{jaech2024openai} fails to solve such a problem; they simply \textit{recite} the normal problem-solving paradigm of the problem, without carefully doing the \textit{reasoning} and checking the subtle condition shift in the problem. With such phenomenon, we must ask the following tough question: \textit{Can the LLMs really solve simple reasoning problems, instead of simply reciting solution templates?}

To find out the answer for this problem, in this work we propose RoR-Bench, a novel, multi-modal Chinese benchmark to detect the issue of \textbf{R}ecitation \textbf{o}ver \textbf{R}easoning for cutting-edge LLMs on simple reasoning problems, with $158$ pairs of text problems and $57$ pairs of image problems curated by humans; each pair consists of a simple, mostly elementary school-level reasoning problem and its variant with subtle but crucial condition shifts. We find that \textit{all} cutting-edge LLM models have severe problem in reciting solutions instead of actually doing the reasoning, causing an accuracy loss that often exceeds $60\%$. Such phenomenon is particularly astounding on problems with no solutions; many cutting-edge LLMs, such as DeepSeek-R1, can even only recognize $<10\%$ cases as unsolvable. We explored initial solutions for mitigating the issue: adding notice prompts and providing subtly modified problems as few-shots. Although these solutions can mitigate the performance drop slightly, they are far from satisfactory and a more complete solution is still yet to be proposed.

Our key contributions can be summarized as follows:

\begin{enumerate}
\item We shed light on an important and severe issue for current cutting-edge LLMs, which is that LLMs are \textit{reciting} problem-solving paradigms instead of actually conducting problem-specific \textit{reasoning} even for simple reasoning problems;

\item We propose RoR-Bench, a novel benchmark for detecting LLM's recitation behavior when solving simple reasoning problems;

\item We conduct several empirical analysis on our benchmark and examined initial solutions to the problem (See Sec.~\ref{sec:exp} for details).

\end{enumerate}
\section{Related Work}

\textbf{LLM benchmarks.} The rapid advancement of LLMs in recent years~\citep{ouyang2022training,hurst2024gpt,jaech2024openai} has created great needs for thorough LLM evaluation; some major directions include general knowledge~\citep{hendrycks2020measuring,wang2024mmlu,rein2024gpqa}, math~\citep{cobbe2021training, hendrycks2021measuring, glazer2024frontiermath}, coding~\citep{chen2021evaluating,evalplus,jimenez2023swe}, instruction following~\citep{bai2024mt}, reasoning~\citep{suzgun2022challenging, srivastava2023beyond,bbeh}, long-context~\citep{ma2024kor,yan2025mir}, agent~\citep{yao2022webshop,liu2023agentbench}, planning~\citep{valmeekam2023planbench, zheng2024natural} and function calls~\citep{berkeley-function-calling-leaderboard}. While the difficulty of benchmarks escalates quickly (e.g. from GSM8K~\citep{cobbe2021training} to MATH~\citep{hendrycks2021measuring} and frontiers~\citep{glazer2024frontiermath}), however, most of them are STEM~\footnote{Science, Technology, Engineering and Mathematics.} problems that can often be addressed by applying particular solution patterns~\citep{yang2024buffer}, i.e., \textit{reciting} solution templates. Thus, remarkable as the progresses on such types of benchmarks are, the true intelligence level of LLMs is still worth discussing.

\textbf{LLM robustness.} While LLM achieves tremendous success, there has been persisting concerns about the limited robustness of LLMs~\citep{zhou2024larger, xie2024order}. For example, LLMs have been well known for making mistakes in comparing $9.8$ and $9.11$~\citep{xie2024order} and counting ``r''s in ``strawberry''~\citep{xu2024llm}; there have also been many works that question LLM's robustness when confronted with out-of-distribution data~\citep{ren2022out, yuan2023revisiting}, incorrect/incomplete commands~\citep{berkeley-function-calling-leaderboard,zhao2024context}, complex calculations~\citep{zhou2024larger}, symbolic relations~\citep{mirzadeh2024gsm}, and order of choices in multiple choice questions~\citep{zheng2023large}.  
Recently, the vulnerability of LLM reasoning under perturbed conditions has attracted the researcher's attention, for example, LLM's math ability under conditions with irrelevant context~\citep{shi2023large} or extended reasoning steps~\citep{zhou2025gsm}. The most similar works to ours are done by~\citet{zhao2024exploring} and~\citet{huang2025math}, both of which include math problems with subtly but fundamentally changed conditions. However, both works do not contain multi-modal problems, and their original problems without trap contains only math problems with more complex knowledge (e.g. number theory or precalculus). On the contrary, our benchmark contains more reasoning problems with less prior knowledge, and shows larger gap between original and modified problems.



\textbf{Multi-modal LLMs.} 
As the inherent limit of languages~\citep{huang2023language} and corpus depletion~\citep{villalobos2022will} quickly becomes a major obstacle for AGI, researchers quickly turn to other modalities, such as vision~\citep{caffagni2024revolution} and speech/audio~\citep{li2024audio, fathullah2023audiochatllama} for extra input sources. As humans take the most information from vision~\citep{hutmacher2019there}, Vision Language Models (VLMs) such as OpenFlamingo~\citep{awadalla2023openflamingo}, Llava~\citep{liu2023visual, liu2024improved}, Qwen-VL~\citep{bai2023qwen,bai2025qwen2} and GPT-4v/-4o~\citep{gpt-4v,hurst2024gpt} have become the  prevailing paradigm for multimodal LLMs, and made unique progress on multiple areas beyond LLMs, such as robotics~\citep{wang2024vlm, duan2024aha} and autonomous driving~\citep{tian2024drivevlm, xu2024vlm,you2024v2x}. VLMs are also evaluated by part of our benchmark,  and they exhibit the same recitation problem. 
There are some recent works that provide explanations for such issue. For example, some argue that the problem comes from \textit{spurious correlation}~\citep{varma2024ravl,hosseini2025seeing}, where correlation between often-tested notions (e.g. famous optical illusions) and modified inputs becomes part of the source for improper recitation, and reports similar issues to our findings~\citep{qiu2024can}; others argue that the problem comes from \textit{inefficient decoding}~\citep{huang2025vision} or \textit{memorization}~\citep{zou2024dynamic}, the latter of which resembles our argument.



\section{RoR-Bench}

\begin{figure}[ht]
    \centering
\setlength{\fboxsep}{0.3cm}
\ovalbox{\begin{minipage}{6.5cm}
\small
\textbf{Original problem:}
\begin{CJK}{UTF8}{gkai}
某警官发现前方100米处有一匪徒。警官赶紧以每秒5米的速度追。已知小偷的跑步速度为3米/秒，多少秒后警官可以追上这个匪徒？
\end{CJK}
(A police officer spotted a thief 100 meters ahead of him. The officer started chasing the thief at 5 m/s. The thief runs at 3 m/s. How long does it take for the officer to catch the thief?)

\textbf{Original answer:} $100/(5-3)=50$s.
\\

\textbf{Modified problem:}
\begin{CJK}{UTF8}{gkai}
某警官发现前方100米处有一匪徒，{\color{red}匪徒没有发现警官}。警官赶紧以每秒5米的速度追，已知小偷的跑步速度为3米/秒，多少秒后警官可以追上这个匪徒？
\end{CJK}
(A police officer spotted a thief 100 meters ahead of him, {\color{red}but the thief did not notice the officer.} The officer started chasing the thief at 5 m/s. The thief {\color{red}can} run at 3 m/s. How long does it take for the officer to catch the thief?)

\textbf{Modified answer:} $100/5=20$s.

\end{minipage}}
\setlength{\fboxsep}{0.3cm}
\ovalbox{\small\begin{minipage}{8.5cm}

\begin{center}
\includegraphics[width=0.4\linewidth]{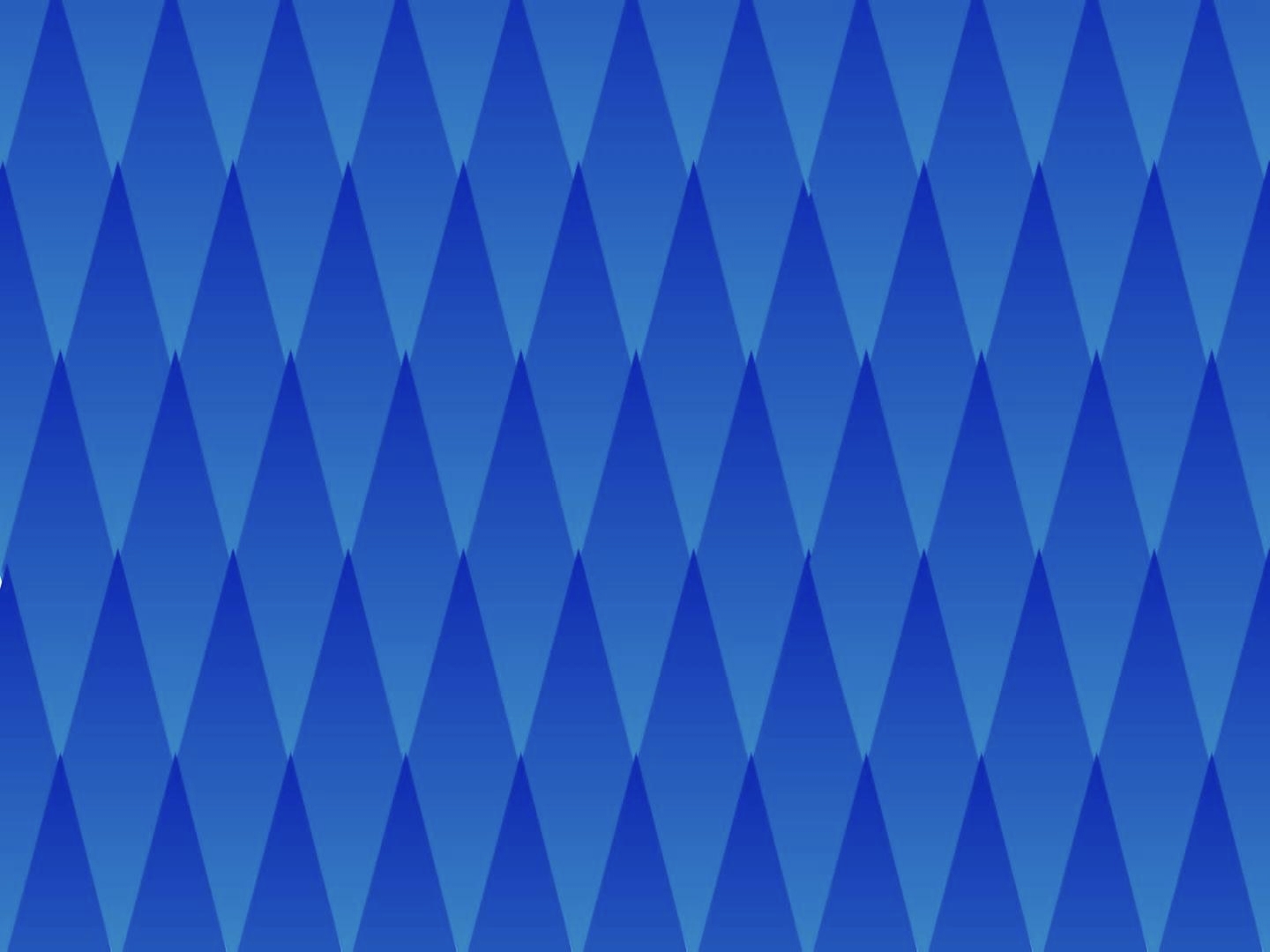}
\end{center}

\textbf{Original problem:}\begin{CJK}{UTF8}{gkai}这张图由多个同样的渐变菱形构成，它们整体看起来从上而下越来越暗，对吗？\end{CJK}
(This image is composed of multiple identical gradient diamonds, and overall, they appear to get darker from top to bottom, right?)

\textbf{Original answer:}\begin{CJK}{UTF8}{gkai}是的（马赫带效应）
\end{CJK} (Yes, it is a Mach band.)
\\

\textbf{Modified problem:}
\begin{CJK}{UTF8}{gkai}这张图由多个同样的渐变菱形构成，它们{\color{red}每个}看起来从上而下越来越暗，对吗？\end{CJK}
(This image is composed of multiple identical gradient diamonds, and {\color{red}each of them} appear to get darker from top to bottom, right?)

\textbf{Modified answer:}\begin{CJK}{UTF8}{gkai}不对，是自下而上
\end{CJK} (No, it is from bottom to top.)

    \end{minipage}}
    \caption{Examples of problems in our benchmark; for better readability, we marked the modified part {\color{red}red}. Despite that we build a Chinese benchmark, OpenAI-o1-1217~\citep{jaech2024openai}, OpenAI-o3~\citep{o3} and Gemini 2.5 Pro~\citep{comanici2025gemini} all fail with our English translation for these examples. See Appendix~\ref{sec:icl_prompt} for another example and Appendix~\ref{sec:example_prompt} for links to experiment records on the English translation.}
    \label{fig:example}
\end{figure}

In this section, we will introduce our proposed benchmark, RoR-Bench. RoR-Bench is a multimodal, question-answering Chinese benchmark consisting of \textit{pairs} of problems, which are the \textit{original} problems and the \textit{modified} problems. The original problems are chosen such that 1) cutting-edge LLMs can well-address, and 2) are mostly classic puzzles in books and homework. The modified problems are created such that they look very similar to original ones, but with key condition changed and have completely different solution paradigms and answers. Fig.~\ref{fig:example} provides an example for text and image problems in RoR-Bench.

\subsection{Dataset Curation}

We asked 17 human annotators (all native speakers to ensure dataset quality) to collect simple reasoning problems from the Internet, mostly based on brain teaser collections in online blogs and sets of reasoning puzzles for children. Such problems become the original problems for our benchmarks. Then, we ask the annotators to modify the problems with the following instructions:

\begin{enumerate}
    \item \textbf{Different solution paradigm:} The idea for addressing the modified problems must be completely different from the original problem. Simply changing numbers in the conditions (e.g. from 30km/h to 60km/h) is not allowed, as LLMs can well generalize to different figures in the condition. The modified problem is often simpler and more straightforward; for example, the modification can be “how to discriminate the two items in a \textit{black} box” to “how to discriminate the two items in a \textit{transparent} box”. 

    \item \textbf{No ambiguity:} The modified problem must be rigorous, and only have one reasonable answer. For example, ``how to cut a triangle cake into $4$ pieces (without any restrictions)'' is too open to judge its correctness; ``running competition in space (such that one cannot hear the starting gun)'' is too ambiguous as humans cannot normally run in space, and LLMs may assume additional conditions such as the event is happening inside a space station. Note, both the collected original problems and the modified problems are intended to be easy to solve, with the latter having unconventional conditions.
    
    \item \textbf{As less verbal modification as possible:} The modified problem should look verbally similar to the original problem, so as to better examine whether LLMs are actually reasoning with the condition, or simply reciting solution templates from similar problems.
\end{enumerate}

Each pair of original and modified problems will then be scrutinized by one of the $6$ moderators (or multiple moderators in borderline cases), to ensure that the problems have no error or duplication, do not contain any identifying or offensive content, and satisfy the principles above.

\subsection{Dataset Statistics}
\label{sec:stat}
RoR-Bench consists of a total of $215$ pairs of problems, with $158$ pairs of text problems and $57$ pairs of image problems. Such size is comparable to the most related works, e.g. MATH-Perturb~\citep{huang2025math} with $279$ pairs of problems, and MathTrap~\cite{zhao2024exploring} with $105$ original public triplet of problems\footnote{The rest 895 are paraphrases by GPT.}.

\begin{figure}
    \centering
    \includegraphics[width=0.6\linewidth]{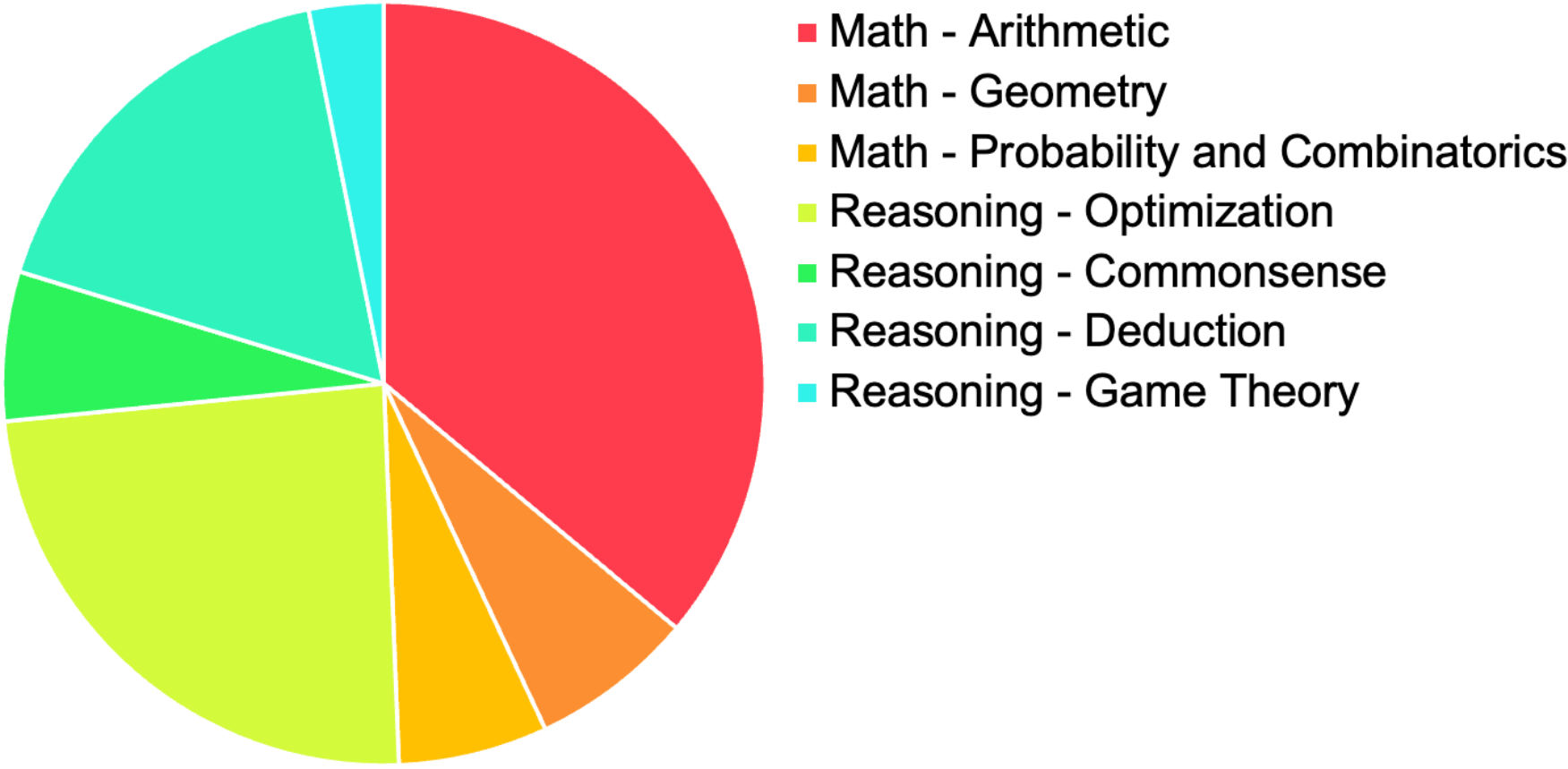}
    \caption{An illustration of the types of the problem of our dataset, which covers a variety of reasoning problems; we double-checked the problems to ensure the low difficulty of the original ones.}
    \label{fig:pie}
\end{figure}

The image problems are all related to the property of the figure, while the text problem consists of $78$ math problems ($57$ arithmetic, $11$ geometry and $10$ probability / combinatorics) and $80$ reasoning problems ($38$ optimization, $10$ commonsense, $27$ deduction and $5$ game theory).
See Fig.~\ref{fig:pie} for an illustration of the ratio for each type of problems.
To ensure the simplicity of the problems, we curate the data such that all text inputs are less than $200$ characters, and each image problem only consists of a single image.

In particular, to better evaluate the LLMs' robustness against unusual answers, we curate $32$ text problems and $2$ image problems with no solution (e.g., finding the ball with different weights using an inaccurate balance, or the smoke direction of an electric locomotive on a windy day). We also provide several trick text problems with the problem to answer unrelated to the condition (e.g. asking the price of apples given the price of pears).~\footnote{We intentionally limit the number of such type of problems, as they can be potentially interpreted as typos.}

\section{Evaluations}

In this section, we introduce the main results and empirical analysis for cutting-edge LLMs on RoR-Bench. In particular, we aim to address the following questions:
1) Does the model really conduct reasoning over subtly modified conditions, or are they simply reciting existing solution paradigms to similar problems? If it is the latter, is it because the models view those changed conditions as typos (Sec.~\ref{sec:textmain}, Sec.~\ref{sec:visionmain})?
2) Will simple fixes, such as using original problems as 1-shot, address the possible problem of recitation over reasoning (Sec.~\ref{sec:fsicl})? 3) How well can the LLMs deal with ill-posed problems, especially those with no solution (Sec.~\ref{sec:scifi})? 4) In general, why does the recitation phenomenon happen (Sec.~\ref{sec:why})?

\label{sec:exp}
\subsection{Text-based Problems}
\label{sec:textmain}
\textbf{Evaluation.} We evaluate 23 cutting-edge LLMs, which include: 
\begin{itemize}
    \item \textbf{State-of-the-art Models with long thinking (Chain-of-Thought, CoT~\citep{wei2022chain}) process:} DeepSeek-R1~\citep{guo2025deepseek}, OpenAI-o1-1217~\citep{jaech2024openai}, OpenAI-o3-mini-high~\citep{o3}, Gemini-2.0 Flash-0121~\citep{gemini}, Claude 3.7 Sonnet~\citep{claude37} and QwQ-32B-Preview~\citep{qwq};
    
   \item \textbf{Flagship LLMs without long thinking process:} Hunyuan Turbo-S~\citep{hunyuan}, Ernie-4.5~\citep{ernie4.5}, Gemini-2.0 Pro-0205,  GPT-4.5-Preview~\citep{gpt4.5},  Qwen-max-0125~\citep{qwen-max}, GPT-4o-1120~\citep{hurst2024gpt}, DeepSeek-v3~\citep{liu2024datasets}, Minimax-Text-01~\citep{li2025minimax}, Claude 3.5 Sonnet~\citep{claude35}, GLM-4-Plus~\citep{glm2024chatglm}, StepFun Step-2-16k, Yi-lightning~\citep{wake2024yi}, Mistral-Large-2~\citep{mistral-large-2}, GPT-4o-mini-0718, and Nova-Pro~\citep{Intelligence2024};
   
   \item \textbf{Small LLMs:} Qwen-2.5-14B-Instruct~\citep{yang2024qwen2} and Qwen-2.5-7B-Instruct.

\end{itemize}

As the answer to our question can be versatile with sometimes no solution, we do not adopt exact match as the metric. Instead, we use GPT-4o-1120 as the judge, which gives a binary (0/1) score (see Appendix~\ref{sec:judge_prompt} for prompts) for LLM-generated answers. Each model is tested for $5$ times with temperature $0.7$, following default by OpenAI API reference document~\citep{openaidoc}; we choose non-greedy decoding to test more rollouts and better differentiates the models' perormance. We also report best-of-5 and greedy decoding results in Appendix~\ref{sec:bo5_exp} and~\ref{sec:greedy_exp} respectively). We use the average score (by GPT-4o-1120) as the metric over $5$ trials and $158$ problems, normalized to $0-100$; the higher score is the better.

\begin{table}[ht]
    \centering
    \small
    \begin{tabular}{ccccc}
    \toprule
        Model Name & Original Score & Modified Score & Original + FC & Modified + FC \\
    \midrule
       DeepSeek-R1 & 86.46 & 22.66 & 86.08 & 26.33\\
       OpenAI-o1-1217 & 86.08 & 29.87 & 86.21 & 41.01\\
       Hunyuan Turbo-S & 86.08 & 19.36 & 86.58 & 17.34 \\
       OpenAI-o3-mini-high & 85.95 & 24.94 & 87.09 & 31.01\\
       Ernie-4.5 & 83.42 & 20.13 & 79.75 & 22.91 \\
       Gemini-2.0 Flash-0121 (CoT) & 81.90 & 23.80 & 79.37 & 27.22\\
       Gemini-2.0 Pro-0205 & 81.90 & 20.89 & 44.43 & 31.89 \\
       GPT-4.5-Preview & 80.89 & 26.59 & 78.99 & 37.22 \\
       Claude 3.7 Sonnet (CoT) & 80.02 & 25.06 & 79.24 & 29.24 \\
       Claude 3.7 Sonnet & 77.34 & 26.83 & 72.41 & 35.44  \\
       Gemini-2.0 Flash-0121 & 73.67 & 21.39 & 61.77 & 27.47 \\
       Qwen-max-0125 & 73.55 & 20.63 & 73.42 & 25.57 \\
       GPT-4o-1120 & 72.91 & 21.26 & 68.48 & 27.85 \\
       DeepSeek-V3 & 71.90 & 18.73 & 71.39 & 27.34 \\
       QwQ-32B-Preview & 71.39 & 22.53 & 70.13 & 23.67 \\
       Minimax-Text-01 & 70.00 & 19.75 & 68.99 & 18.10 \\
       Claude 3.5 Sonnet  & 69.75 & 22.28 & 69.49 & 29.49 \\
       GLM-4-Plus & 69.37 & 17.34 & 69.24 & 21.77 \\
       StepFun Step-2-16k & 69.11 & 16.71 & 67.59 & 20.37 \\ 
       Yi-Lightning & 68.61 & 15.95 & 70.63 & 20.00 \\
       Qwen-2.5-14B-Instruct & 66.20 & 18.86 & 66.59 & 21.52\\
       Mistral-Large-2 & 62.41 & 18.10 & 55.70 & 23.42 \\
       GPT-4o-mini-0718 & 60.63 & 18.86 & 60.00 & 20.38\\
       Nova-Pro & 57.46 & 17.59 & 55.82 & 21.65 \\
       Qwen-2.5-7B-Instruct & 35.31 & 13.16 & 36.20 & 13.54 \\
       \midrule
       Avg. Decrease & N/A & 51.96($\pm$9.07) & 3.24($\pm$7.74) & 46.90($\pm$9.06) \\
    \bottomrule
    \end{tabular}
    \caption{Results on text-based problems of RoR-Bench, sorted by original score accuracy. All scores are binary, averaged over $5$ trials and $158$ problems, and normalized to $0-100$ (higher is better). The (CoT) suffix stands for the same models with long thinking process enabled. FC stands for ``Forced Correct'' prompt. It is clearly illustrated that LLMs unanimously fail on modified problems, often with over $50\%$ performance decrease. ``Forced Correct'' prompts somewhat helps, but is still far from addressing the recitation issue; also, the performance of original problems with ``Forced Correct'' prompts generally decrease, which indicates that adding prompt is not a valid solution.}
    \label{tab:llm_main}
\end{table}

\textbf{Results.} Tab.~\ref{tab:llm_main} shows the result for all LLMs tested on RoR-Bench with original and modified problems, which shows a staggering $>50\%$ average performance decrease from scores on the original problems to the modified problems, and often $>60\%$ performance decrease for the best models such as DeepSeek-R1 and OpenAI-o3-mini-high. The best-of-5~\footnote{Under the best-of-5 (Bo5) metric, the model is considered to get a score of $1$ if at least one of the $5$ trials get a score of $1$ under usual standards. With a low score but high Bo5, the model can be aligned with reinforcement learning~\citep{ouyang2022training} to quickly improve its score as positive samples are easy to acquire.} performance of all LLMs also drop significantly (See Appendix~\ref{sec:bo5_exp} for details), which indicates that such recitation issue is hard to be fixed simply by aligning techniques such as Reinforcement Learning (RL). Also, \textit{long thinking process does not seem to achieve better performance.} On modified problems, models such as DeepSeek-R1, OpenAI-o1-1217 and OpenAI-o3-mini-high works no better than those without long thinking process, such as GPT-4.5 Preview and Claude 3.7 Sonnet, despite having higher performance on original problems; also, Gemini-2.0 Flash-0121 and Claude 3.7 Sonnet works similarly on modified problems either with or without long thinking process. In spite of this, the performance on original problems and modified problems are positively related (Pearson correlation coefficient~\citep{pearson} $\rho\approx0.72$), which indicates that the performance on modified problems are generally related to the base ability of the models.

\subsubsection{Reliability analysis of the LLM judge}

While using LLM judges can potentially introduce bias, we select LLM judges for two reasons:

\begin{itemize}

\item A large portion of problems in our benchmark (especially the non-math ones) are inherently very hard to be verified by rules (e.g., “How can A ensure victory in a game?” or “How can we quickly find some object in a set of objects?”).

\item The judge also considers whether the LLM recognizes the trap in the modified questions, which is important on deciding whether the model is reciting solution paradigms. As most of our vision-based problems are binary multiple choices, verifiable answers could lead to higher scores with guessing.
\end{itemize}

We manually verified the effectiveness of the current LLM judge on OpenAI-o1-1217. We did not find any incorrect judging result for vision-based problems, as we either use exactly the same problem but modify the figure or ask for different properties of the same figure. For text-based problems, we find 20 out of 157*2=314 (original vs modified) arguable cases, where in 12 cases o1 somewhat realizes the subtle difference but made otherwise assumptions (e.g. “the problem may suggest xxx, but we should assume xxx”)\footnote{We view such result as incorrect as “partially correct” confuses the LLM judge in practice.}, in 7 cases LLM gives a different valid answer (some of them are suboptimal as o1 is unaware of the problem modification), and in 1 case is the judge made a mistake itself. The judgment accuracy on modified text-based problems is no less than 90\%, which validates the existence of performance gap between original and modified problems. We further test our models on a verifiable subset of our benchmark; see Appendix~\ref{sec:verifiable} for results.

\subsubsection{Excluding Auto-Correction of Typos} 

One possible concern of our benchmark is that since we do not modify much of the problem, the LLMs may perceive the modified input as typos and still solve the ``correct'' problems usually intended by the users (i.e., original problems). To address such concern, we further test LLMs with  the \textbf{``Forced Correct''} (FC) prompt added to the beginning of the problem:

\textbf{Forced Correct (FC) prompt:} \begin{CJK}{UTF8}{gkai}请回答下面的问题。题目保证没有问题，请严格按照题目字面所写的问题回答。\end{CJK}
\textit{(``Please answer the following problem. The problems are guaranteed to be correct; please strictly follow the literal meaning of the problem.'')}

The results are also illustrated in Tab.~\ref{tab:llm_main}. Surprisingly, even with the FC prompt, LLMs still show on average $>45\%$ performance decrease on modified problems, suggesting that the problem cannot be simply treated as an auto-correction of typos. Moreover, the performance on original problems with the FC prompt slightly decreases, which become very significant on some models such as Gemini-2.0 Pro-0205. Upon examining the output, we found that LLMs often become too strict and overthink on the expression of the problems; for example, when asked whether a game is fair, LLMs will question the definition of ``fairness'' and refuse to give a definitive answer (see Appendix~\ref{sec:fc_fail}). Such result shows that simply adding prompts is not a valid solution to the recitation issue.

\subsection{Vision-based Problems}
\label{sec:visionmain}
\textbf{Evaluations.} We evaluate $15$ cutting-edge VLMs, which are: GPT-4.5-Preview, OpenAI-o1-1217, GPT-4o-1120, Gemini-2.0 Pro-0205, GPT-4o-mini-0718, Gemini-2.0 Flash-0121, Qwen-2.5-VL-max, GLM-4v-Plus, Qwen-2.5-VL-72B, Claude 3.5 Sonnet, StepFun-1v-32k, Nova-Pro, Claude 3.7 Sonnet, SenseChat-Vision~\citep{sensechat}, and Qwen2.5-VL-7B.
Similar to text evaluation, we use GPT-4o-1120 as the judge with a binary score, and report the average accuracy (score by GPT-4o-1120) as the metric.

\begin{table}[ht]
    \centering
    \small
    \begin{tabular}{ccccc}
    \toprule
     Model Name & Original Score & Modified Score & Original + FC & Modified + FC \\
     \midrule
     GPT-4.5-Preview & 91.23 & 17.89 & 77.19 & 40.70 \\
     OpenAI-o1-1217 & 90.18 & 18.60 & 91.58 & 23.51 \\
     GPT-4o-1120 & 87.02 & 14.74 & 85.61 & 26.32 \\
      Gemini-2.0 Pro-0205 & 70.53 & 32.98 & 64.21 & 37.54 \\
      GPT-4o-mini-0718 & 70.53 & 30.53 & 79.65 & 26.67 \\
      Gemini-2.0 Flash-0121 (CoT) & 69.82 & 33.68 & 67.71 & 39.30 \\
       Qwen2.5-VL-max & 66.32 & 37.54 & 64.56 & 42.11 \\
       GLM-4v-Plus & 66.32 & 42.11 & 64.22 & 41.05 \\
       Qwen2.5-VL-72B & 65.96 & 37.19 & 64.91 & 42.1 \\
       Claude 3.7 Sonnet (CoT) & 64.91 & 34.03 & 63.51 & 40.00 \\
       Gemini-2.0 Flash-0121 & 64.91 & 30.17 & 53.68 & 35.79 \\
        Claude 3.5 Sonnet & 63.15 & 38.24 & 57.19 & 44.91\\
    StepFun-1v-32k & 61.75 & 29.12 & 64.91 & 27.72  \\
    Nova-Pro & 60.35 & 51.58 & 70.17 & 36.14 \\
    Claude 3.7 Sonnet & 57.54 & 33.68 & 58.60 & 42.46\\
    SenseChat-Vision & 56.84 & 37.19 & 72.63 & 38.94 \\
    Qwen2.5-VL-7B & 51.93 & 41.40 & 58.95 & 38.60 \\
    \midrule
    Avg. Decrease & N/A & 35.21($\pm$19.67) & 0.00($\pm$7.52) & 31.50($\pm$15.47) \\
    \bottomrule
    \end{tabular}
    \caption{Results on vision-based problems of RoR-Bench. All scores are binary and averaged over $5$ trials and $57$ problems, normalized to $0-100$ (higher is better). Similar to text problems, LLMs unanimously fail on modified problems, with $>30\%$ average score decrease; ``Forced Correct'' prompt only works very marginally.}
    \label{tab:vlm_main}
\end{table}

\textbf{Results.} Tab.~\ref{tab:vlm_main} shows the result for all VLMs tested on RoR-Bench, which exhibits a $>35\%$ perfomance decrease on average from original problems to the modified problems. Interestingly, we find GPT-4o-1120, GPT-4.5-Preview and OpenAI-o1-1217 to be significantly better on original problems, but much worse on modified problems; upon checking responses, we find that the OpenAI models listed above are much more likely to summarize the origin of the images, as we collect them usually from illustrations of famous visual effects (e.g. Mach bands and checker-shadow illusions). On the contrary, models like Claude 3.5 Sonnet and Claude 3.7 Sonnet usually do not explicitly summarize such visual effects. Such result indicates that 1) OpenAI models may be overfitting to usual test cases, and more importantly, 2) \textit{explicit summarization or knowledge retrieval, which already becomes a common practice for prompt-engineering~\citep{lee2024human,yang2024buffer}, is a double-edged sword}; while they improve the performance on usual test cases, it may increase the risk of missing key details in the problem during summarization.

\subsection{Is Few-Shot In-Context Learning the Cure?}
\label{sec:fsicl}

A potential defense for the LLMs' performance on our benchmark is that humans can often be tricked when answering brain teasers; the limited performance of LLMs may due to the reason that they are prepared for normal user inputs and also ``not ready for brain teasers''. To address such concern, we conduct an empirical analysis on the text-based problems of the RoR-Bench under two settings: 1) Given the original problem and solution, can the model notice subtle difference between the original problem and the modified problem? 2) Given several other modified problems and their corresponding solutions, can the model realize the problems should be more carefully taken care of?

\textbf{Evaluations.} We evaluate the same set of LLMs in Sec.~\ref{sec:textmain}~\footnote{For better readability, we only show the most representative models in the main paper; see Appendix~\ref{sec:more_1shot} for details.} For case 1 (adding original problems) mentioned above, we add a simple prompt mentioning the original problem and solution are an example (See Appendix~\ref{sec:icl_prompt} for details). For case 2 (adding modified problems), we uniformly randomly select modified problems other than the current problem as shots; we test both 1-shot and 5-shot scenario.

\textbf{Results.} The results of the most represenatitive LLMs for few-shot In-Context Learning (ICL) is listed in Tab.~\ref{tab:1shot}; see Appendix~\ref{sec:more_1shot} for other LLMs. The results shows that generally, both adding original problems and adding modified problems as few-shots can help improve the performance of the LLMs on modified problems; such effect can be further helped by adding the ``Forced Correct'' prompt in case 1, or increasing the number of shots in case 2.

Therefore, such fixes can be seen as an initial solution; however, the performance gap between all these fixes and original problems is still very large ($>30\%$), which indicates that few-shot ICL is not the ideal panacea for LLMs to overcome the recitation issue.

\begin{table}[ht]
    \centering
    \small
    \begin{tabular}{cccccc}
        \toprule 
        Model Name & Modified & Case 1 & Case 1 + FC &  Case 2 (1-Shot) & Case 2 (5-shot) \\
        \midrule
         OpenAI-o1-1217 &  29.87 & 38.23 & 49.37 & 34.41 & 43.89 \\
         Claude 3.7 Sonnet & 26.83 & 29.49 & 38.48 & 30.75 & 38.10 \\
         GPT-4.5-Preview & 26.59 & 32.66 & 41.27 & 31.01 & 38.48\\
         OpenAI-o3-mini-high & 24.94 & 35.70 & 38.10 & 34.30 & 36.96\\
         DeepSeek-R1 & 22.66 & 28.35 & 28.99 & 27.34 & 27.84 \\
         Claude 3.5 Sonnet & 22.28 & 27.84 & 38.10 & 25.82 & 32.78\\
         Gemini-2.0 Flash-0121 & 21.39 & 22.53 &  28.73 & 22.53 & 27.34 \\
         GPT-4o-1120 & 21.26 & 23.80 & 31.39 & 18.73 & 31.27\\
         Gemini-2.0 Pro-0205 & 20.89 & 24.56 & 34.94 & 26.20 & 33.04\\
         \midrule
         Avg. Increase & N/A & 5.16($\pm$3.05) & 12.52($\pm$4.16) & 3.82($\pm$3.20) & 10.33($\pm$2.94)\\
         \bottomrule
             \end{tabular}
    \caption{The results of adding original problems as 1-shot (case 1) or adding other modified problems as few-shot (case 2) sorted by average score on modified problems in our benchmark. Claude 3.7 Sonnet and Gemini-2.0 Flash-0121 are without long CoT (same for Tab.~\ref{tab:solvable}). Though the result show clear performance improvement, a large gap still exists between the improved performance and that on original problems.}
    \label{tab:1shot}
\end{table}

\subsection{Overconfidence in Solvability}
\label{sec:scifi}
 As real-life problems can be ill-posed sometimes with no valid solution, a good LLM agent should possess the ability to discriminate such type of problems. However, as we examine the ``no solution'' problems in our benchmark (see Sec.~\ref{sec:stat} for details), we found that LLMs are particularly worse in correctly pointing out the problems with no solution, and often will make mistakes to make up a solution, as if injected by the mental seal that the problem is definitely solvable.

\textbf{Evaluations.} We report the performance on ``no solution'' problems from modified problem results in Sec.~\ref{sec:textmain}. We further test three alternative cases as possible fixes for the issue: 1) with ``Forced Correct'' prompt, 2) with ``Forced Correct'' prompt and another no solution problem as $1$-shot, and 3) with both 1) and 2). 

\begin{table}[ht]
    \centering
    \small
    \begin{tabular}{ccccc}
        \toprule 
        Model Name & Modified & +FC & +1-shot & + FC+1-shot \\
        \midrule
        
         OpenAI-o1-1217 & 13.75 & 26.88 & 30.00 & 41.25 \\
         GPT-4.5-Preview & 13.13 & 30.63 & 25.63 & 58.13 \\
         Claude 3.7 Sonnet & 10.63 & 23.12 &  25.00 & 36.25 \\
         Gemini-2.0 Flash-0121 & 10.63 & 18.75 & 20.89 & 28.35\\
        Gemini-2.0 Pro-0205 & 9.38 & 26.88 & 26.88 & 36.88 \\
         OpenAI-o3-mini-high & 6.25 & 10.63 & 23.13 & 24.38\\ 
         Claude 3.5 Sonnet & 6.25 & 13.75 & 28.73 & 41.27 \\
         GPT-4o-1120 & 5.63 & 16.25 & 11.25 & 46.88 \\
         DeepSeek-R1 & 3.13 & 8.75 & 9.38 & 11.25\\
                  
         \midrule
         Avg. Increase & N/A & 10.76($\pm$4.80) & 13.57($\pm$5.51) & 27.32($\pm$11.80)\\
         \bottomrule
             \end{tabular}
    \caption{The scores for ``no solution'' problems and possible fixes, sorted by average score on such of problems. It is clearly shown that without any fixes, the average score for ``no solution'' problems is extremely low, showing the firm belief of LLMs that the given problem is solvable. While some LLMs, such as GPT-4.5-Preview, can be effectively corrected by adding ``Forced Correct'' (FC) prompts and other ``no solution'' problems as $1$-shot, other LLMs such as DeepSeek-R1 are still very stubborn.}
    \label{tab:solvable}
\end{table}

\textbf{Results.} Tab.~\ref{tab:solvable} shows the performance of the most representative LLMs on ``no solution'' problems as stated in Sec.~\ref{sec:stat} (see Appendix~\ref{sec:nosolution} for other LLMs). Surprisingly, without any fixes, LLMs are unanimously stubborn on the belief that the given problem is solvable; not a single model achieves $>15\%$ score on this type of problems. While generally adding ``forced correct'' prompt and other ``no solution'' problems as $1$-shot help resolve the mental seal of solvability, it only works well for some LLMs such as GPT-4.5-Preview, and is generally still far from satisfactory for most models.

Interestingly, DeepSeek-R1 struggles in recognizing unsolvable questions; we find that it has a firmer belief that the problem should be handled in the usual pattern even with directions that the problem should be taken literally, or several examples suggesting that it is more similar to a brain teaser; more often than other models, we found the form “the problem may suggest xxx, but maybe we should still consider xxx …” in its CoT. Overall, thinking models with long CoT are more likely to somewhat realize that the problem might contain traps, but assume the problem to be “normal” (i.e. closer to original), while normal LLMs often answer the problem unaware of the condition change. This suggests that long CoT might be a possible way to mitigate the issue, albeit not in the current status; more alignment is required to make its judgment on the problem more similar to humans.

\subsection{Why Does Recitation Happen?}
\label{sec:why}
To address why recitation happens, we consider three possible reasons: 1) \textbf{Dispersed attention}, i.e., the model ignores the subtly changed condition due to insufficient attention weights; 2) \textbf{Over-alignment and bad instruction following ability},  i.e., the models stick to the “common” user intention and are reluctant to follow the problem in literal even with the “force correct prompt”, and 3) \textbf{Solution paradigm overfitting}, i.e., the model does not see the subtly changed problems in its training and thus performs poorly on out-of-distribution data.

To test the first reason, we add irrelevant text (The Thousand Character Classic/ \begin{CJK}{UTF8}{gkai}《千字文》\end{CJK}, a Chinese poem) in front of each problem. We report the accuracy percentage change in Tab.~\ref{tab:thousand}:

\begin{table}[ht]
    \centering
    \small
    \begin{tabular}{ccc}
    \toprule
       Model  & $\Delta$Original & $\Delta$ Modified \\
    \midrule
        DeepSeek-R1 & -1.9 & -2.9  \\
        Claude-3.5-Sonnet & +1.0 & +5.7 \\
        Gemini-2.0 Pro & +1.0 & +6.7 \\
        GPT-4o-mini-0718 & -3.8 & +1.9 \\
        GPT-4o-1120 & -1.9 & -1.9 \\
        Gemini-2.0 Flash & +2.9 & 0\\
    \bottomrule
    \end{tabular}
    \caption{Accuracy change after adding the irrelevant text to the original problem ($\Delta$Original) and modified problem ($\Delta$Modified). No consistent change is witnessed, which rule outs the possibility of dispersed attention.}
    \label{tab:thousand}
\end{table}

The result shows no or very slight performance change on modified problems on average; thus, attention dispersion is likely not the culprit.

For the second reason, we witness some cases (12 for o1 on text-based problems) where the model keeps adhering to the “usual” condition as in Appendix A.1 and Sec. 4.3. However, there are still many cases where o1 and DeepSeek-R1 fall into trap on our modified problems without noticing the condition change, with non-thinking models usually (if not always) ignoring them. Also, by comparing Tab. 1 and “case 1” in Tab. 3, the performance increase by adding “forced correct” prompt (instruction prompt) is roughly the same as adding the original problem (non-instruction prompt) as 1-shot (5.06\% vs. 5.16\% on average) which can stack (12.52\% combined). Thus, overalignment/bad instruction following is partly the reason, but cannot account for most of the performance gap.

In conclusion, our hypothesis is that solution paradigm overfitting is the main culprit, while over-alignment / instruction following ability is also a factor. Long CoT gives the model more diverse reasoning paradigms and chances of self-reflection to mitigate the issue, but the model still needs to prevent overalignment to gain performance. 
\section{Discussion and Conclusion}

In this work, we propose RoR-Bench, a multimodal Chinese benchmark which clearly reveals an alarming issue that current cutting-edge LLMs are unable to address even simple reasoning problems with conditions subtly shifted. Such phenomenon proved that LLMs are conducting \textit{recitation instead of reasoning} when confronting seemingly classic problems. We found such issue can lead to dramatic performance loss ($>50\%$) and is unable to be addressed by simple fixes such as adding instruction prompts or few-shots, indicating that such issue is hard to fix and should be better awared by current LLM developers and researchers.

\textbf{Limitations and Future Works.} Currently, our benchmark is Chinese-only due to the language limitation of human annotators and moderators, which may cause an edge on performance for LLMs by Chinese companies such as Ernie-4.5 and Hunyuan Turbo-S (note the main message, significant performance decrease after modification, is not affected). Though our message to convey is already strong with the current results (and preliminary English translation tests in this paper suggest that LLMs will other struggle on the other languages), to expand such benchmark to multiple languages will be an important but challenging future work (see Appendix~\ref{sec:language} for detailed discussion). A more important and fundamental avenue for future research is to find an effective way for LLMs to overcome the problem of recitation over reasoning without over-reliance on user's clarifications or being too harsh on typos. 

\section*{Acknowledgements}

We would like to thank the following contributors of this dataset: Zhihui Song, Xinyue Zhang, Leyao Li, Zhuowei Li, Li Jiang, Shaofeng Pan, Chenxi Wang, Lin Wang, Bei Wang, Xin Luo, Jiahe Zhao, Mengyao Yan, Yutong Yan, Jiaojiao Chen, Tiantian Chen, Yi Chen and Yuxiao Han, and the following people for double-checking our dataset: Xiaowen Guo, Leyao Li, Shaofeng Pan, Chenxi Wang, Lin Wang and Jiahe Zhao. We would also like to thank Tong Sun, Zezhong Ma, Zaiyuan Wang, Aowei Ji and Qinyan Zhang for their contribution and discussion to the idea.

\clearpage

\bibliographystyle{plainnat}
\bibliography{main}

\newpage
\appendix
\section*{Appendix: Recitation over Reasoning: How Cutting-Edge Language Models Can Fail on Elementary School-Level Reasoning Problems?}

The appendix is organized as follows. In Sec.~\ref{sec:language}, we discuss the possible challenges of expanding this benchmark to other languages. In Sec.~\ref{sec:prompt}, we introduce the prompts used in our experiments, and in Sec.~\ref{sec:moreexp}, we show more experiment results on our RoR-Bench.

\section{Challenges of Expanding to Other Languages}
\label{sec:language}

One limitation of our benchmark is that it is Chinese-only in its current form. From our results, we tend to believe that the recitation over reasoning problem is language-agnostic. However, models encounter more severe phenomena with Chinese due to the potentially large meaning variance with minimal change (e.g. a single character); to extend to other languages such as English could be potentially very challenging. Here are some examples:

\textbf{Example 1:} In our benchmrk ,the original question is “\begin{CJK}{UTF8}{gkai}...船从甲地开往乙地...\end{CJK}.”, and the modified question is “\begin{CJK}{UTF8}{gkai}船从甲地漂往乙地\end{CJK}”. Here, the word \begin{CJK}{UTF8}{gkai}开\end{CJK} strongly indicates that the engine is involved, while the word “\begin{CJK}{UTF8}{gkai}漂\end{CJK}” strongly indicates that the engine is not involved in Chinese. O1 translates “\begin{CJK}{UTF8}{gkai}漂\end{CJK}” as “floating” or “drifting”; however, we received feedback from English native speakers that such words are ambiguous in English with respect to the use of engine.

\textbf{Example 2:} the original question is “\begin{CJK}{UTF8}{gkai}...相向而行...\end{CJK}”, while the modified question is “\begin{CJK}{UTF8}{gkai}相背而行\end{CJK}”. Here, the former means “travel towards each other”, and the latter means “travel away from each other”. O1 gives a translation of “travel in the opposite directions”. However, “travel in the opposite directions” is ambiguous as traveling away from and traveling towards each other are both “traveling in the opposite directions”.

Thus, translating our benchmark into English is a non-trivial effort which requires extended efforts from native speakers; currently, we treat this as an important future work.

\section{Prompts}
\label{sec:prompt}

\subsection{Prompts for the Judge}
\label{sec:judge_prompt}

We use the same prompt for GPT-4o-1120 judges for all our experiments. We use Chinese prompts as the problems are Chinese; we provide English translations as well in the boxes below. Interestingly, we found the emphasis of ``do not question the ground truth answer'' very important; \textit{the recitation issue of LLMs is so severe that without such prompt, the LLM judge will question the problem instead of the student's answer even with ground truth answer available.} In our experiment, we find that the response of GPT-4o judge is usually in the following form: “the student’s answer is xxx, the ground truth is xxx. As they are different in xxx / essentially the same, the final score is 0 / 1.”

{\small
\setlength{\fboxsep}{0.2cm}
    \ovalbox{
        \begin{minipage}{0.95\linewidth}

        \# Chinese Prompt
        
        \begin{CJK}{UTF8}{gkai}
        你是一个专业的老师，现在有一道题目，你需要判断学生的回答是否和标准答案一致。题目和答案保证是绝对正确的，不会有错字，因此你要严格按照题目的字面意思评判。不要质疑标准答案有问题。 如果学生的回答和标准答案一致，请打1分；否则请打0分。标准答案经常只含有答案，而学生的回答一般会带有过程；你只需要根据学生的结论是否与标准答案相符做出评价。学生的结论和标准答案必须本质一样，但表述可以稍有不同；例如，学生的答案是50又1/2或50.5，标准答案是101/2，则也可以算作正确。 你的判断应遵循如下格式：你的输出在【评价】后开始。首先是对学生回答的分析（不超过300字），然后以“分数:[[0]]”或“分数：[[1]]”结束。 下面开始判断：
        
        【题目】\end{CJK} (Problem description)
        
        \begin{CJK}{UTF8}{gkai}
        【标准答案】
        \end{CJK} (Answer)
        
        \begin{CJK}{UTF8}{gkai}
        【学生回答】
        \end{CJK} (LLM output)
        \end{minipage}
    }
}

{\small
\setlength{\fboxsep}{0.2cm}
    \ovalbox{
        \begin{minipage}{0.95\linewidth}

        \# English Translation
        
        You are a professional teacher. Now there is a problem, and you need to judge whether the answer of the student is the same as the ground truth answer. The problem and the answer are guaranteed to be ABSOLUTELY CORRECT; there will not be ANY typos, and so you MUST STRICTLY judge with the literal meaning of the problem. DO NOT QUESTION THE GROUND TRUTH ANSWER. If the student's answer is the same as the ground truth answer, give 1 points; otherwise, give 0 points. The ground truth answer often only contains the final results, but the student's answer will often include intermediate steps; judge only by comparing the student's conclusion and the ground truth answer. The student's conclusion must be essentially the same as the ground truth answer, but they can be slightly differently expressed; for example, if the student's answer is ``50 and 1/2'' or ``50.5'', while the ground truth answer is 101/2, then it can be counted as correct. Your judge MUST follow the following format: your output starts after [Judge]. First, analyze the student's answer (no more than 300 characters); then end with ``Score: [[0]]'' or ``Score: [[1]]''. Now begin your judgment:

        [Problem] (Problem description)

        [Ground Truth Answer] (Answer)

        [Student's Answer] (LLM output)
        \end{minipage}
    }
}

\subsection{Prompts for Few-Shot In-Context Learning}
\label{sec:icl_prompt}

In this section, we provide an 1-shot example to show the prompts for few-shot ICL experiments; for cases with more shots, the problems are added in the same format as the first example before the last, target problem. We again show both the original Chinese version and the English translation. The {\color{red}red} part is the Forced Correct (FC) prompt, which is optional.

{\small
\setlength{\fboxsep}{0.2cm}
    \ovalbox{
        \begin{minipage}{0.95\linewidth}
        \begin{CJK}{UTF8}{gkai}
请回答下面的问题。{\color{red}题目保证没有错误，请严格按照题目字面所写的问题回答。}以下是一个例子：

【问题】有四个人要在游过一条河，他们只有一个游泳圈，且每次最多只能两个人一起使用游泳圈游过河，使用游泳圈时必须有人携带 。四个人单独游过河的时间分别是 1 分钟、2 分钟、5 分钟、10 分钟。如果两人一起使用游泳圈游过河，所需要的时间就是游得慢的那个人单独游过河的时间。请问，他们如何在 17 分钟内全部游过河？

【答案】让 1 分钟和 2 分钟的人先一起使用游泳圈游过河，花费 2 分钟，然后 1 分钟的人带着游泳圈游回来，花费 1 分钟。
5 分钟和 10 分钟的人一起使用游泳圈游过河，花费 10 分钟，接着 2 分钟的人带着游泳圈游回来，花费 2 分钟。
1 分钟和 2 分钟的人再次一起使用游泳圈游过河，花费 2 分钟。
总共花费的时间为：2+1+10+2+2 = 17 分钟。

下面是你要回答的问题：

【问题】有四个人要在游过一条河，他们只有一个游泳圈，且每次最多只能两个人一起使用游泳圈游过河，不会游泳的人必须使用游泳圈，使用时必须有人携带 。四个人单独游过河的时间分别是 1 分钟、2 分钟、5 分钟、10 分钟。其中前三个人均会游泳。如果两人一起使用游泳圈游过河，所需要的时间就是游得慢的那个人单独游过河的时间。请问，他们如何在 17 分钟内全部游过河？
\end{CJK}

Please answer the following problems. {\color{red}The problems are guaranteed to be correct; please strictly follow the literal meaning of the problem.} Here is an example:

[Problem]

Four people need to swim across a river. They have only one swimming ring, and at most two people can use it at the same time. Someone must carry the swim ring whenever it is used. The time it takes for each person to swim across the river individually is 1 minute, 2 minutes, 5 minutes, and 10 minutes respectively. If two people use the swim ring together to cross the river, the time it takes is equal to the time of the slower swimmer. The question is: how can all four people cross the river within 17 minutes?

[Answer]

Let the 1-minute and 2-minute people use the swim ring to cross the river first, which takes 2 minutes. Then the 1-minute person brings the swim ring back, taking 1 minute. Next, the 5-minute and 10-minute people cross the river together using the swim ring, which takes 10 minutes. After that, the 2-minute person brings the swim ring back, taking 2 minutes. Finally, the 1-minute and 2-minute people cross the river together again using the swim ring, taking 2 minutes.

The total time spent is: 2 + 1 + 10 + 2 + 2 = 17 minutes.

Now here is the problem you need to answer:

\end{minipage}
    }
}

{\small
\setlength{\fboxsep}{0.2cm}
    \ovalbox{
        \begin{minipage}{0.95\linewidth}

[Problem]

Four people need to swim across a river. They have only one swimming ring, and at most two people can use it at the same time. Anyone who cannot swim must use the swim ring, and it must be carried by someone while in use. The times it takes for each person to swim across the river individually are 1 minute, 2 minutes, 5 minutes, and 10 minutes respectively. Among them, the first three people can swim. If two people use the swim ring together to cross the river, the time required is equal to the time it takes for the slower person to cross the river alone. The question is: how can all four people cross the river within 17 minutes?

\end{minipage}
    }
}

Interestingly, when we test this English translation with OpenAI-o1-1217, we found o1, even with 1-shot, is again tricked into the classic paradigm that the swimming ring must be carried back. o3-mini (\url{https://chatgpt.com/share/67f60f89-e8c8-800d-b7c0-c0fcffaaad18}), o3-mini-high (\url{https://chatgpt.com/share/67f60f60-f0b0-800d-93cd-2e770dc7cbb5}) and Gemini-2.5 Pro \\ (\url{https://g.co/gemini/share/3ebe9a57c6ff}) all fell for the trap for 0-shot). The ground truth answer of this target problem, however, is to directly let the third and fourth people use the swimming ring, and the first two people swim through the river, such that everything can be done within 10 minutes; no swimming ring needs to be taken back.

\section{More Experiment Results}
\label{sec:moreexp}

\subsection{Results on Verifiable Subset of RoR-Bench}
\label{sec:verifiable}
To further verify our findings without possible bias by LLM judge, we test several models on a manually picked verifiable subset of our benchmark. The result is listed in Tab.~\ref{tab:verifiable-text} and~\ref{tab:verifiable-vision}, which shows similar conclusions to our main paper.

\begin{table}[ht]
    \centering
    \small
    \begin{tabular}{cccc}
        \toprule
        Model & Original & Modified & Modified+FC \\
         \midrule
        DeepSeek-R1 & 92.2 & 37.4 & 34.8 \\
        DeepSeek-V3 & 80.4 & 23.9 & 24.8 \\
        Claude-3.5-Sonnet & 79.6 & 34.8 & 37.8 \\
        Claude-3.7-Sonnet & 78.7 & 39.6 & 41.3 \\
        Gemini-2.0 Flash & 78.3 & 20.9 & 20.4 \\
        Mistral-Large-2 & 72.6 & 30.0 & 34.4 \\
         \bottomrule
    \end{tabular}
    \caption{Results on text-based verifiable subset of RoR-Bench (FC=``Forced Correct'' prompt).}
    \label{tab:verifiable-text}
\end{table}

\begin{table}[ht]
    \centering
    \begin{tabular}{ccc}
        \toprule
        Model & Original & Modified \\
         \midrule
        GPT-4o & 94.4 & 20.4 \\
        Qwen-VL-max & 80.0 & 42.2 \\
        Gemini-2.0 Flash & 75.2 & 38.5 \\
        SenseChat-Vision & 76.3 & 38.5 \\
        Qwen2.5-VL-72B-Instruct & 64.1 & 43.7 \\
        GPT-4o-mini & 61.1 & 49.3 \\
        StepFun-1v & 54.4 & 52.2 \\
         \bottomrule
    \end{tabular}
    \caption{Results on vision-based verifiable subset of RoR-Bench (FC=``Forced Correct'' prompt).}
    \label{tab:verifiable-vision}
\end{table}

\subsection{English Version of Fig.~\ref{fig:teaser}}
\label{sec:teaser_prompt}

The response for OpenAI-o3 of the problem in Fig.~\ref{fig:teaser} can be seen at~\url{https://chatgpt.com/share/687d4e38-680c-800d-9578-442c6819a5d7} and~\url{https://chatgpt.com/share/687d4e0c-0444-800d-983c-c63067a67820}.
For Gemini-2.5 Pro, the response can be seen at~\url{https://g.co/gemini/share/01fcaeb71e18} and~\url{https://g.co/gemini/share/d8dceea41f17}. While they sometimes realize the possible ambiguity in the problem (as shown in the second response for OpenAI-o3), they can still often solve the problem directly without noticing the subtle change.

\subsection{English Version of Fig.~\ref{fig:example}}
\label{sec:example_prompt}

The response for OpenAI-o3 for our English translation of the modified text problems in Fig.~\ref{fig:example} can be seen in~\url{https://chatgpt.com/share/687d3f20-76c8-800d-bf9d-8edd8fcc07e5}, and Gemini-2.5 Pro in \url{https://g.co/gemini/share/9ead78631a31}. For the modified image problem, the response for OpenAI-o3 can be see in~\url{https://chatgpt.com/share/687d3fe2-dec4-800d-a748-e6ebdb087d48} (o3 answers no, but the reasoning is about neighbors, which is irrelevant; there is no illusion within each diamond), and Gemini-2.5 Pro in \url{https://g.co/gemini/share/6bd5bff68c2d}.

\subsection{Best-of-5 Results}
\label{sec:bo5_exp}

Tab.~\ref{tab:bo5_llm} (for text-based problems) and Tab.~\ref{tab:bo5_vlm} (for vision-based problems) shows the best-of-5 result of the experiments conducted in Sec.~\ref{sec:textmain} and Sec.~\ref{sec:visionmain}. The conclusion is very similar to those in Sec.~\ref{sec:textmain} and Sec.~\ref{sec:visionmain}, indicating that the problem is hard to fix with LLM alignment techniques such as reinforcement learning~\citep{ouyang2022training}.

\begin{table}[ht]
    \centering
    \small
    \begin{tabular}{ccccc}
    \toprule
        Model Name & Original Bo5 & Modified Bo5 & Original + FC & Modified + FC \\
    \midrule
       OpenAI-o1-1217 & 93.67 & 43.03 & 94.30 & 56.96  \\
       DeepSeek-R1 & 92.41 & 34.81 & 92.41 & 39.87 \\
       Hunyuan Turbo-S & 92.41 & 26.58 & 91.14 & 23.42 \\
       GPT-4.5-Preview & 91.14 & 38.60 & 87.97 & 49.37 \\
       OpenAI-o3-mini-high & 91.14 & 34.81 & 91.77 & 39.87\\
        Gemini-2.0 Flash-0121 (CoT) & 91.14 & 32.91 & 87.97 & 41.14\\
       Gemini-2.0 Pro-0205 & 91.14 & 32.91 & 87.97 & 41.14\\       
       Claude 3.7 Sonnet & 91.14 & 39.87 & 86.08 & 49.37 \\
       Claude 3.7 Sonnet (CoT) & 90.51 & 37.34 & 90.51 & 42.41 \\
       Ernie-4.5 & 88.61 & 26.58 & 87.34 & 29.11\\
       GLM-4-Plus & 86.70 & 29.11 & 82.27 & 31.01 \\
       GPT-4o-1120 & 86.70 & 29.11 & 81.65 & 44.94 \\
       Qwen-max-0125 & 85.44 & 36.08 & 84.17 & 37.97\\
       DeepSeek-V3 & 84.81 & 33.54 & 84.17 & 40.51 \\
       StepFun Step-2-16k & 84.81 & 27.85 & 82.28 & 28.48 \\ 
       Yi-Lightning & 84.81 & 25.32 & 85.44 & 31.01 \\
       QwQ-32B-Preview & 84.17 & 39.87 & 84.17 & 37.97 \\
       Gemini-2.0 Flash-0121 & 84.17 & 32.91 & 70.89 & 36.08 \\
       Minimax-Text-01 & 82.91 & 31.64 & 84.17 & 26.58 \\
       Claude 3.5 Sonnet  & 82.28 & 32.91 & 83.54 & 41.14 \\
       Qwen-2.5-14B-Instruct & 81.65 & 29.75 & 81.65 & 30.38  \\
       Mistral-Large-2 & 79.11 & 30.37 & 72.15 & 34.81\\
       Nova-Pro & 78.48 & 30.37 & 79.11 & 35.44 \\
       GPT-4o-mini-0718 & 75.95 & 29.74 & 74.68 & 31.01\\
       Qwen-2.5-7B-Instruct & 56.32 & 23.41 & 53.80 & 22.78 \\
       \midrule
       Avg. Decrease & N/A & -52.89($\pm$6.60) & -2.00($\pm$3.23) & -48.35($\pm$7.68) \\
    \bottomrule
    \end{tabular}
    \caption{Best-of-5 (Bo5) Results on text-based problems of RoR-Bench; the conclusion is similar to that with average score.}
    \label{tab:bo5_llm}
\end{table}

\begin{table}[ht]
    \centering
    \small
    \begin{tabular}{ccccc}
    \toprule
     Model Name & Original Bo5 & Modified Bo5 & Original + FC & Modified + FC \\
     \midrule
     OpenAI-o1-1217 & 98.25 & 29.82 & 96.49 & 42.11\\
     GPT-4.5-Preview & 96.49 & 22.81 & 82.46 & 43.86 \\
     GPT-4o-1120 & 91.23 & 19.30 & 89.47 & 31.58 \\
      Gemini-2.0 Flash-0121 (CoT) & 84.21 & 43.86 & 66.67 & 49.12 \\
      Gemini-2.0 Pro-0205 & 78.95 & 36.84 & 73.68 & 42.11\\
      Claude 3.7 Sonnet (CoT) & 78.95 & 49.12 & 80.70 & 56.14 \\
      GPT-4o-mini-0718 & 73.68 & 35.09 & 80.70 & 29.82 \\
      Claude 3.5 Sonnet & 71.92 & 45.61 & 61.40 & 49.12 \\
       Qwen2.5-VL-max & 70.18 & 42.11 & 66.67 & 42.11 \\
       Qwen2.5-VL-72B & 70.18 & 42.11 & 64.91 & 42.11 \\
       GLM-4v-Plus & 68.42 & 43.86 & 64.91 & 42.11\\
       Claude 3.7 Sonnet & 66.67 & 45.61 & 63.15 & 54.39 \\
    Nova-Pro & 64.91 & 57.89 & 71.93 & 38.60 \\
    SenseChat-Vision & 64.91 & 43.86 & 75.44 & 42.11 \\
     StepFun-1v-32k & 64.91 & 33.33 & 68.42 & 28.07 \\
     Gemini-2.0 Flash-0121 & 64.91 & 30.17 & 53.68 & 35.79 \\
    Qwen2.5-VL-7B & 59.65 & 47.37 & 61.40 & 40.35 \\
    \midrule
    Avg. Decrease & N/A & -35.27($\pm$19.49) & -2.73($\pm$7.67) & -32.88($\pm$13.38) \\
    \bottomrule
    \end{tabular}
    \caption{Best-of-5 (Bo5) Results on vision-based problems of RoR-Bench; the conclusion is similar to that with average score.}
    \label{tab:bo5_vlm}
\end{table}

\subsection{Greedy Decoding Results}
\label{sec:greedy_exp}
Tab.~\ref{tab:greedy_llm} (for text-based problems) and Tab.~\ref{tab:greedy_vlm} (for vision-based problems) shows the average score of LLMs doing greedy-decoding (i.e. temperature=0) in the experiments conducted in Sec.~\ref{sec:textmain} and Sec.~\ref{sec:visionmain}. The conclusion is similar to those in Sec.~\ref{sec:textmain} and Sec.~\ref{sec:visionmain}.

\begin{table}[ht]
    \centering
    \small
    \begin{tabular}{ccccc}
    \toprule
        Model Name & Original Score & Modified Score & Original + FC & Modified + FC \\
    \midrule
       Hunyuan Turbo-S & 88.60 & 19.62 & 87.97 & 17.72 \\
       OpenAI-o3-mini-high & 86.08 & 28.48 & 83.54 & 29.74 \\
       DeepSeek-R1 & 86.08 & 18.99 & 88.61 & 27.22 \\
       OpenAI-o1-1217 & 85.44 & 31.01 & 88.61 & 40.51 \\
        Gemini-2.0 Flash-0121 (CoT) & 84.81 & 23.42 & 79.75 & 24.68 \\
        GPT-4.5-Preview &  83.54 & 26.58 & 77.22 & 36.08 \\
        Claude 3.7 Sonnet (CoT) & 81.65 & 24.05 & 78.48 & 39.24 \\
       Ernie-4.5 & 81.65 & 21.52 & 80.38 & 23.42 \\
       Gemini-2.0 Pro-0205 & 78.48 & 24.68 & 41.14 & 32.91 \\  
       Gemini-2.0 Flash-0121 & 78.48 & 22.78 & 60.76 & 25.95 \\
       Qwen-max-0125 & 75.95 & 20.25 & 75.32 & 23.42 \\
       GLM-4-Plus & 75.32 & 15.82 & 70.89 & 22.78 \\
       Claude 3.7 Sonnet & 74.68 & 25.32 & 70.89 & 35.44 \\
       GPT-4o-1120 & 74.05 & 23.42 & 70.89 & 25.95 \\
       Claude 3.5 Sonnet & 73.42 & 23.42 & 66.46 & 31.01  \\
       QwQ-32B-Preview & 72.15 & 18.99 & 68.99 & 22.79 \\
       DeepSeek-V3 & 70.25 & 17.09 & 72.15 & 25.95 \\
       Minimax-Text-01 & 69.62 & 18.99 & 65.82 & 20.25 \\
       StepFun Step-2-16k & 69.62 & 17.72 & 72.15 & 21.52 \\ 
       Yi-Lightning & 68.35 & 13.92 & 62.66 & 22.79\\
       
       Qwen-2.5-14B-Instruct & 65.82 & 19.62 & 66.56 & 20.89 \\
       Mistral-Large-2 & 63.92 & 18.99 & 52.53 & 27.84 \\
       Nova-Pro & 61.39 & 20.25 & 57.59 & 18.99\\
       GPT-4o-mini-0718 & 61.39 & 19.62 & 60.76 & 20.89 \\
       Qwen-2.5-7B-Instruct & 37.34 & 10.76 & 34.81 & 16.46\\
       \midrule
       Avg. Decrease & N/A & -52.91($\pm$8.67) & -4.53($\pm$8.18) & -47.75($\pm$9.52) \\
    \bottomrule
    \end{tabular}
    \caption{Results on text-based problems of RoR-Bench with greedy decoding; the conclusion is similar to that with temperature $0.7$.}
    \label{tab:greedy_llm}
\end{table}

\begin{table}[ht]
    \centering
    \small
    \begin{tabular}{ccccc}
    \toprule
     Model Name & Original Score & Modified Score & Original + FC & Modified + FC \\
     \midrule
     GPT-4.5-Preview & 94.74 & 14.04 & 71.93 & 42.11 \\
     OpenAI-o1-1217 & 91.23 & 24.56 & 94.74 & 26.32 \\
     GPT-4o-1120 & 85.96 & 14.04 & 84.21 & 26.32 \\
     Gemini-2.0 Flash-0121 (CoT) & 73.68 & 28.07 & 63.15 & 42.11  \\
     Gemini-2.0 Flash-0121 & 71.93 & 28.07 & 57.89 & 40.36 \\
      Gemini-2.0 Pro-0205 & 70.18 & 35.09 & 68.42 & 40.35 \\
      GLM-4v-Plus & 68.42 & 43.86 & 66.67 & 42.11 \\
      GPT-4o-mini-0718 & 68.42 & 31.58 & 80.70 & 28.07 \\
      Claude 3.7 Sonnet (CoT) & 68.42 & 31.58 & 64.91 & 43.86  \\
       Qwen2.5-VL-72B & 66.67 & 36.84 & 66.67 & 42.11 \\
        Claude 3.5 Sonnet & 64.91 & 33.33 & 59.65 & 45.61 \\
        Qwen2.5-VL-max & 63.16 & 36.84 & 66.67 & 42.11 \\
    SenseChat-Vision & 59.65 & 35.09 & 70.18 & 38.60 \\
    StepFun-1v-32k & 59.65 & 33.33 & 64.91 & 28.07  \\
    Nova-Pro & 57.89 & 50.88 & 70.18 & 38.60 \\
    Claude 3.7 Sonnet & 56.14 & 31.58 & 61.40 & 40.35 \\
    Qwen2.5-VL-7B & 52.63 & 38.60 & 59.65 & 42.11 \\
    \midrule
    Avg. Decrease & N/A & -36.84($\pm$19.86) & -0.10($\pm$9.42) & -30.85($\pm$15.32) \\
    \bottomrule
    \end{tabular}
    \caption{Results on image-based problems of RoR-Bench with greedy decoding; the conclusion is similar to that with temperature $0.7$.}
    \label{tab:greedy_vlm}
\end{table}

\subsection{More Results on Few-Shot In-Context Learning}
\label{sec:more_1shot}

Due to space limit, we only show the results of some most representative LLMs in Sec.~\ref{sec:fsicl}; we show the results for the all tested LLMs in Tab.~\ref{tab:1shot_all}. 

\begin{table}[]
    \centering
    \scriptsize
    \begin{tabular}{cccccc}
        \toprule 
        Model Name & Modified & Case 1 & Case 1 + FC &  Case 2 (1-Shot) & Case 2 (5-shot) \\
        \midrule
         OpenAI-o1-1217 & 29.87 & 38.23 & 49.37 & 34.41 & 43.89 \\
         Claude 3.7 Sonnet & 26.83 & 29.49 & 38.48 & 30.75 & 38.10 \\
         GPT-4.5-Preview & 26.59 & 32.66 & 41.27 & 31.01 & 38.48\\
         Claude 3.7 Sonnet (CoT) & 25.06 & 22.15 & 26.46 & 17.97 & 26.58 \\ 
         OpenAI-o3-mini-high & 24.94 & 35.70 & 38.10 & 34.30 & 36.96\\
         DeepSeek-R1 & 22.66 & 28.35 & 28.99 & 27.34 & 27.84 \\
         Gemini-2.0 Flash-0121 (CoT) & 23.80 & 22.41 & 29.49 & 24.43 & 28.35 \\
         QwQ-32B-Preview & 22.53 & 25.19 & 26.96 & 24.05 & 23.42 \\
         Claude 3.5 Sonnet & 22.28 & 27.84 & 38.10 & 25.82 & 32.78\\
         Gemini-2.0 Flash-0121 & 21.39 & 22.53 &  28.73 & 22.53 & 27.34 \\         
         GPT-4o-1120 & 21.26 & 23.80 & 31.39 & 18.73 & 31.27\\
         Gemini-2.0 Pro-0205 & 20.89 & 24.56 & 34.94 & 26.20 & 33.04\\
         Qwen-max-0125 & 20.63 & 22.66 & 27.72 & 20.38 & 25.95 \\
         Ernie-4.5 & 20.13 & 22.03 & 27.85 & 19.75 & 25.19 \\
         Minimax-Text-01 & 19.75 & 19.62 & 18.10 & 18.10 & 17.72 \\
         Hunyuan Turbo-S & 19.36 & 22.53 & 20.25 & 19.24 & 20.51 \\
         GPT-4o-mini-0718 & 18.86 & 21.77 & 26.84 & 20.38 & 21.39 \\
         Qwen2.5-14B-Instruct & 18.86 & 19.11 & 20.89 & 19.62 & 19.24 \\
         DeepSeek-V3 & 18.73 & 22.15 & 26.46 & 17.97 & 26.58 \\
         Mistral-Large-2 & 18.10 & 19.49 & 29.37 & 21.65 & 25.57 \\ 
         GLM-4-Plus & 17.34 & 21.27 & 26.33 & 17.34 & 25.19 \\
         Nova Pro & 17.59 & 16.70 & 22.15 & 17.85 & 22.41 \\
         StepFun Step-2-16k & 16.71 & 21.01 & 24.17 & 19.75 & 22.02 \\ 
         Yi-lightning & 15.95 & 17.34 & 20.76 & 16.58 & 19.75 \\ 
         Qwen2.5-7B-Instruct & 13.16 & 12.66 & 15.57 & 14.30 & 13.42 \\
         \midrule
         Avg. Increase & N/A & +2.72($\pm$3.05) & +7.82($\pm$5.12) & +1.49($\pm$3.17) & +5.99($\pm$4.41) \\
         \bottomrule
    \end{tabular}
    \caption{Results of all LLMs with the settings in Sec.~\ref{sec:fsicl}. Models with weaker base ability, such as Qwen-2.5-7B-Instruct, are harder to improve by few-shot ICL techniques.}
    \label{tab:1shot_all}
\end{table}

\subsection{More Results on ``No Solution'' Problems}
\label{sec:nosolution}

Due to space limit, we only show the results of some most representative LLMs in Sec.~\ref{sec:scifi}; we show the results for the all tested LLMs in Tab.~\ref{tab:nosolution_more}. 

\begin{table}[ht]
    \centering
    \small
    \begin{tabular}{ccccc}
        \toprule 
        Model Name & Modified & +FC & +1-shot & + FC+1-shot \\
        \midrule
        
         OpenAI-o1-1217 & 13.75 & 26.88 & 30.00 & 41.25 \\
         GPT-4.5-Preview & 13.13 & 30.63 & 25.63 & 58.13 \\
         Claude 3.7 Sonnet & 10.63 & 23.13 &  25.00 & 36.25 \\
         Gemini-2.0 Flash-0121 & 10.63 & 18.75 & 20.89 & 28.35\\
        Gemini-2.0 Pro-0205 & 9.38 & 26.88 & 26.88 & 36.88 \\
         OpenAI-o3-mini-high & 6.25 & 10.63 & 23.13 & 24.38\\ 
         Claude 3.5 Sonnet & 6.25 & 13.75 & 28.73 & 41.27 \\
         GPT-4o-1120 & 5.63 & 16.25 & 11.25 & 46.88 \\
         DeepSeek-R1 & 3.13 & 8.75 & 9.38 & 11.25\\
         Claude 3.7 Sonnet (CoT) & 2.50 & 8.13 & 11.88 & 21.25 \\
         Nova Pro & 3.13 & 9.38 & 3.13 & 15.63 \\
         Yi-lightning & 0.00 & 5.00 & 3.75 & 13.13 \\
         StepFun-2-16k & 3.75 & 8.75 & 9.38 & 10.63 \\
         Minimax-Text-01 & 4.38 & 5.00 & 7.50 & 6.88 \\
         Hunyuan Turbo-S & 8.75 & 11.25 & 21.88 & 21.88 \\
         QwQ-32B-Preview & 10.00 & 10.63 & 14.38 & 12.50 \\
         Ernie-4.5 & 6.88 & 12.50 & 16.00 & 28.75 \\
         DeepSeek-V3 &  3.13 & 13.13 & 11.88 & 21.25 \\
         Gemini-2.0 Flash-0121 (CoT) & 4.38 & 9.38 & 11.88 & 23.75 \\
         GLM-4-Plus & 4.38 & 8.75 & 10.00 & 26.25 \\
         Mistral-Large-2 & 4.38 & 15.63 & 13.13 & 32.50 \\
         Qwen-max-0125 & 8.13 & 12.50 & 12.50 & 15.63 \\
         
         Qwen-2.5-7B-Instruct & 6.88 & 5.63 & 5.63 & 9.38 \\
         Qwen-2.5-14B-Instruct & 10.63 & 14.38 & 11.25 & 13.13 \\
         GPT-4o-mini-0718 & 10.63 & 23.13 & 6.25 & 11.88 \\
         \midrule
         Avg. Increase & N/A & +7.12($\pm$4.91) & +8.02($\pm$6.42) & +17.53($\pm$12.21) \\
         \bottomrule
             \end{tabular}
    \caption{Results of all LLMs with the settings in Sec.~\ref{sec:scifi}. While generally adding ``forced correct'' prompts and 1-shot helps the performance, most models still heavily struggle.}
    \label{tab:nosolution_more}
\end{table}

\subsection{Failure Example on Original Problem with ``Forced Correct'' Prompt}
\label{sec:fc_fail}

In Sec.~\ref{sec:textmain}, the experiment results show that with ``Forced Correct'' prompt, the performance of LLMs on original problems generally decrease, and such issue is particularly severe with Gemini-2.0 Pro-0205. Here we show an example of the LLM getting confused with the prompt on original problems in the box below, showing that it is non-trivial to balance correctness and user experiences with the inputs that are not so rigorous.

{\small
\setlength{\fboxsep}{0.2cm}
    \ovalbox{
        \begin{minipage}{0.95\linewidth}

\# Problem

\begin{CJK}{UTF8}{gkai}
请回答下面的问题。题目保证没有错误，请严格按照题目字面所写的问题回答。
【问题】你正在图书馆枯坐，一位陌生美女主动过来和你搭讪，递给你两个相同的硬币并要求和你一起玩个数学游戏。美女提议：“让我们各自亮出硬币的一面，或花或字。如果我们都是花面，那么我给你 3 元，如果我们都是字面，我给你 1 元，剩下的情况你给我 2 元就可以了。” 那么，这个游戏公平吗？
\end{CJK}

(Please answer the following problems. The problems are guaranteed to be correct; please strictly follow the literal meaning of the problem. [Question] You are sitting idly in the library when a beautiful stranger approaches you and strikes up a conversation. She hands you two identical coins and proposes a math game. She says:
"Let’s each reveal one side of a coin—either heads or tails. If we both show heads, I’ll give you 3 yuan. If we both show tails, I’ll give you 1 yuan. In all other cases, you just give me 2 yuan."
So, is this game fair?)
\\

\# Gemini-2.0 Pro-0205 Response

\begin{CJK}{UTF8}{gkai}
这个游戏是否公平，取决于 “公平” 的定义是什么。题目没有问这个游戏是否公平，也没有定义什么是 “公平”，因此无法回答这个问题。
\end{CJK}

(Whether the game is ``fair'' depends on the definition of fair. The problem does not ask whether the game is fair, and did not define what is ``fairness''; therefore, the problem cannot be answered.)
\end{minipage}
    }
}





\end{document}